\documentclass[sigconf,screen]{acmart} 
\pdfoutput=1

\setcopyright{acmlicensed}
\acmPrice{15.00}
\acmDOI{10.1145/3597926.3598045}
\acmYear{2023}
\copyrightyear{2023}
\acmSubmissionID{issta23main-p54-p}
\acmISBN{979-8-4007-0221-1/23/07}
\acmConference[ISSTA '23]{Proceedings of the 32nd ACM SIGSOFT International Symposium on Software Testing and Analysis}{July 17--21, 2023}{Seattle, WA, United States}
\acmBooktitle{Proceedings of the 32nd ACM SIGSOFT International Symposium on Software Testing and Analysis (ISSTA '23), July 17--21, 2023, Seattle, WA, United States}
\received{2023-02-16}
\received[accepted]{2023-05-03}

\usepackage{todonotes}
\usepackage{comment}
\usepackage{graphicx}
\usepackage[T1]{fontenc}
\usepackage{listings}
\usepackage{xcolor}
\definecolor{verylightgray}{rgb}{.97,.97,.97}

\def\inline{\lstinline[basicstyle=\normalsize\ttfamily,deletekeywords={Is,is}]}

\usepackage{wrapfig}
\usepackage{booktabs}
\usepackage{subcaption}
\usepackage{amsmath}
\usepackage{hyperref}[]
\usepackage{tikz}
\usepackage[]{caption}
\usepackage{multicol}

\usetikzlibrary{shapes, shapes.symbols, patterns, arrows, shapes, positioning,
	calc, fit, shadows, automata, shapes.geometric, decorations.text}
\usepackage[outline]{contour}
\contourlength{4pt}
\usepackage{multirow}
\def\inline{\lstinline[basicstyle=\normalsize\ttfamily,deletekeywords={Is,is}]}

\usepackage{pgfplots,pgfplotstable}
\usepgfplotslibrary{statistics}
\pgfplotsset{compat=1.8}

\usepackage{algorithm}
\usepackage[noend]{algpseudocode}
\algrenewcommand\Return{\State \algorithmicreturn{} }%

\algdef{SE}[DOWHILE]{Do}{doWhile}{\algorithmicdo}[1]{\algorithmicwhile\ #1}
\algnewcommand{\True}{\textbf{true}}
\algnewcommand{\False}{\textbf{false}}
\algnewcommand\algorithmicforeach{\textbf{for each}}
\algdef{S}[FOR]{ForEach}[1]{\algorithmicforeach\ #1\ \algorithmicdo}

\usepackage{soul}
\definecolor{darkgreen}{rgb}{0.0, 0.5, 0.0}
\newcommand{\rev}[1]{#1}

\newcommand{\cam}[1]{#1}


\begin{document}

\title{Semantic-Based Neural Network Repair}

\author{Richard Schumi}

\affiliation{%
	\institution{Singapore Management University}
	\country{Singapore}}
\email{rschumi@smu.edu.sg}

\author{Jun Sun}
\affiliation{%
	\institution{Singapore Management University}
	\country{Singapore}}
\email{junsun@smu.edu.sg}

\begin{abstract}
Recently, neural networks have spread into numerous fields including many safety-critical systems. Neural networks are built (and trained) by programming in frameworks such as TensorFlow and PyTorch. Developers apply a rich set of pre-defined layers to manually program neural networks or to automatically generate them (e.g., through AutoML). Composing neural networks with different layers is error-prone due to the non-trivial constraints that must be satisfied in order to use those layers. In this work, we propose an approach to automatically repair erroneous neural networks. The challenge is in identifying a minimal modification to the network so that it becomes valid. Modifying a layer might have cascading effects on subsequent layers and thus our approach must search recursively to identify a ''globally'' minimal modification. Our approach is based on an executable semantics of deep learning layers and focuses on four kinds of errors which are common in practice. We evaluate our approach for two usage scenarios, i.e., repairing automatically generated neural networks and manually written ones suffering from common model bugs. The results show that we are able to repair 100\% of a set of randomly generated neural networks (which are produced with an existing AI framework testing approach) effectively and efficiently (with an average repair time of 21.08s) and 93.75\% of a collection of real neural network bugs (with an average time of 3min 40s).
\end{abstract}


\maketitle

\section{Introduction}
Artificial intelligence (AI) is based on imitating natural intelligence or learning behaviour with a machine. The method called deep learning (DL) employs artificial neural networks to simulate the neurons of brains \cite{DBLP:journals/nn/Schmidhuber15}. 
In recent years, DL has advanced into numerous fields, like language processing, face and speech recognition, due to improvements in computer hardware~\cite{DBLP:journals/csur/PouyanfarSYTTRS19}. 

Although AI systems have been successfully used in many domains, they are not failure-free.
Especially in safety critical applications, like autonomous driving, or medical systems, even minor bugs can have severe consequences. 
Studies of bugs in AI systems
\cite{DBLP:conf/icse/IslamPNR20,DBLP:conf/issta/ZhangCCXZ18,DBLP:conf/issre/ThungWLJ12} have analysed various posted bug reports (or issues), and the large number of identified posts illustrates the high frequency of such bugs.
These studies evaluated the time it took from posting a bug until it was resolved and the time ranges from weeks to months depending on the type of the issue.

Moreover, there are other factors that make it especially difficult and time-consuming to debug AI systems. Neural networks can have a long training time of days or weeks, which makes it cumbersome to evaluate potential ways of fixing a neural network.
Often the error messages that occur during AI development can be unrelated to the actual issue that needs to be fixed \cite{DBLP:conf/icse/IslamPNR20}, the error messages can be inconsistent, or there can even be hidden issues that do not produce error messages \cite{exais}.
Another difficulty comes from the underlying AI frameworks such as TensorFlow or PyTorch. These frameworks are still being rapidly developed, and
a new release may not be backward compatible and can break existing code \cite{DBLP:conf/icse/IslamPNR20}. Furthermore, the provided documentation can sometimes be vague or buggy \cite{li2020documentation}, which makes it hard for developers to use unfamiliar AI components.

In order to make it easier for developers to fix and prevent such bugs, it is important to be able to systematically (and automatically) test and repair the underlying neural networks or deep learning models.
Thus, in this work, we present a novel semantic-based neural network repair approach that helps AI developers fix common errors in the architecture and with the parameters of AI models.

Designing AI or deep learning models is a cumbersome task that requires a lot of expertise, domain knowledge, and effort. AI frameworks provide dozens of deep learning layers that can perform simple mathematical functions or complex operations, like different convolutions or recurrent layers. 
Most of these layers have preconditions, e.g., for the input data or for the parameters. Building a valid model requires the fulfilment of all the preconditions from all layers of the model, which can be challenging since layers can not only be combined in sequences, but in general can be in the form of directed acyclic graphs.
Given such a graph, every connection can cause potential precondition violations, and identifying them can be cumbersome. Moreover, the debug information that is provided by AI frameworks can be imprecise or inconsistent \cite{exais}.
Due to that, the process of developing a new model can be time-consuming and frustrating. 
With our approach, an AI developer is provided with automatically generated changes that can repair the model. The changes are presented in a list, which is ordered based on a change indicator value that reflects the number (and magnitude) of changes that are required. Our approach facilitates and speeds up the time-consuming task of finding a valid model architecture and consistent model parameters. This can be especially helpful for new AI developers, who still lack the necessary skills to fix such issues on their own.
Moreover, our approach can support other AI techniques, like AutoML~\cite{automl} where, e.g., a model architecture can be automatically derived for a given problem, or automatic AI model generation \cite{DBLP:conf/sigsoft/WangYCLZ20}, which can be used in various scenarios, like AI framework testing. For both these techniques, it can be hard to find valid model architectures, especially if randomness is involved. 
To avoid generating invalid models, existing AutoML or AI framework testing approach often rely on a limited set of models defined by templates.
By repairing generated models, unnecessary pruning of the search space can be avoided, which might benefit both these techniques.

Our approach relies on an existing semantics called ExAIS~\cite{exais} which defines the functionality of almost all TensorFlow layers in the logical programming language Prolog. ExAIS contains a number of preconditions that produce debug messages and enable the identification of model bugs. Based on these messages, we apply multiple algorithms to fix different types of errors. For example, for a dimension error (which occurs when the input does not have the required number of dimensions), a simple fix strategy could be to introduce a reshape layer. 

However, it is usually not as easy to repair a neural network with just a single change, because in most cases there need to be further modifications in the following layers as a consequence, i.e., structural changes like that can also cause invalid weight shapes deeper in the network. Moreover, there are usually multiple potential fixes for a bug, and it can be  challenging to find the best repair in different scenarios.
\cam{There is already an existing tool called} Tensfa~\cite{DBLP:conf/issre/WuSC0Q21,DBLP:journals/infsof/WuSCJQ22} \cam{that can automatically repair dimension errors or more generally tensor shape faults, which are also bugs that we are targeting with our method, but in contrast our approach can also repair different types of bugs and it works more straight forward as we will explain later.}

To sum up, the main contributions of our work are:
\begin{itemize}
\item We present a novel semantic-based AI model repair approach that is able to suggest multiple model changes that can fix an invalid AI model.
\item We collect a set of real bugs from AI developers in order to  evaluate our approach in a realistic setting.
\item Additionally, we performed an evaluation with randomly generated models in order to test our approach for a diverse set of models and bugs.
\end{itemize}

\paragraph{Structure.} The rest of the paper is structured as follows. In Sect.~\ref{sec:background}, we introduce necessary background information, e.g., about the semantics that we apply.
In Sect.~\ref{sec:method}, we present the underlying algorithms of our repair approach in detail.
In Sect.~\ref{sec:evaluation}, we show an evaluation with two usage scenarios.
Lastly, we review the related work in Sect.~\ref{sec:related} and conclude in Sect.~\ref{sec:conclusion}.

\section{Background} \label{sec:background}


In this section, we describe the technologies that will be supported by our repair approach as well as the underlying semantics.

\subsection{AI Framework Testing}
While our primary goal is to support programmers when they are developing AI models manually, another interesting application that can be supported by our approach is AI framework (such as TensorFlow and PyTorch) testing. 
%
There is an active line of research on automatic testing of AI frameworks~\cite{DBLP:conf/sigsoft/WangYCLZ20,DBLP:conf/icse/PhamLQT19,li2020documentation,DBLP:conf/aaai/SrisakaokulWAAX18,DBLP:conf/icse/DingKH17,DBLP:conf/issta/DwarakanathASRB18,DBLP:conf/icst/SharmaW19}. These works are motivated by the fact that
bugs in AI frameworks might potentially affect all AI applications that are built with such frameworks. Due to that and also due to the fact that AI frameworks can still suffer from severe bugs \cite{exais}, it is important to thoroughly and systematically test them.
Existing AI framework testing techniques can be categorized into the following groups:
differential testing~\cite{DBLP:conf/icse/PhamLQT19,DBLP:conf/aaai/SrisakaokulWAAX18,DBLP:conf/sigsoft/WangYCLZ20} and metamorphic testing~\cite{DBLP:conf/icse/DingKH17,DBLP:conf/issta/DwarakanathASRB18,DBLP:conf/icst/SharmaW19}.
Differential testing is a technique that compares multiple frameworks (or implementations) against each other, and bugs are discovered when there is an inconsistency. 
Metamorphic testing for AI frameworks usually introduces changes in the input data for the AI model that should not affect the output, and a different output (prediction result) would suggest an issue in the AI framework.
For both these techniques, it is important to have a wide variety of valid AI  models
in order to test different aspects of the AI frameworks. 

Our repair approach can support the generation of such models by enabling effective random model generation that would normally produce a high percentage of invalid models or by allowing other approaches to explore more complicated scenarios that might often be dismissed due to difficulties in finding a working model.

\subsection{AutoML}
Another application that can be supported by our repair approach is 
Automatic machine learning (AutoML) \cite{automl,automl1}.
AutoML, which is also called neural architecture search (NAS) in the deep learning context, is a term for methods that try to reduce the need for manual model building for various learning tasks, like image/object recognition, or language modelling. 
Developing AI systems usually requires a lot of expertise and effort, e.g., for data pre-processing tasks, like feature engineering and to find a deep learning model, i.e., to choose a model architecture and to tune the model in order to achieve an acceptable accuracy. 
There has been a lot of progress with AutoML techniques, which can potentially take over such tasks and sometimes produce models with higher accuracies compared to hand-crafted models from experienced AI developers \cite{DBLP:conf/iclr/ZophL17}.
There are many AutoML approaches that try to find model architectures and tune parameters with different strategies.

Depending on the search strategy, some AutoML approaches can produce invalid models that have to be discarded \cite{DBLP:conf/iclr/ZophL17} which hinders the performance of AutoML. To avoid generating a large number of invalid models, AutoML approaches usually apply only a limited set of predefined layers or groups of layers (a.k.a. cells) to build neural networks based on knowledge
from manually developed networks, which might hinder the discovery of architectures that outperform existing ones \cite{DBLP:journals/jmlr/ElskenMH19}. 
With our approach, we can support a variety of layers and neural network architectures. Hence, we believe that our semantic-based repair method can  help to enable AutoML approaches to explore an extended search space.

\subsection{ExAIS}
Our technique utilises an executable AI semantics called ExAIS~\cite{exais} that is written in the logical programming language Prolog~\cite{nilsson1990logic}.
Prolog is a declarative language that relies on first order logic. 
Programs in Prolog are built with rules and facts, which are usually concerned with mathematical relations. The declarative nature of the language facilitates the creation of  high level semantics. Moreover, it supports various list operations and mathematical expressions, which makes it convenient for specifying deep learning layer behaviour that is often concerned with high dimensional input and mathematical operations.

Listing~\ref{lst:dense} shows the Prolog semantics of a Dense layer~\cite{dense} of ExAIS. It is a standard densely connected layer, which contains a number of nodes, each of which is connected with all inputs. 
The output is computed as the sum of all inputs (multiplied with the weights) at each node with an added bias value.
%
The specification works as follows. The rule starting in Line~1 has multiple parameters: a list \inline{[I|Is]}, a weight array \inline|IWs|, and a bias array \inline|Bs| (which can be intuitively regarded as inputs) and a list \inline|Os| (which can be regarded as the output). The list notation \inline{[I|Is]} enables access to the first element \inline{I} and the remaining elements \inline{Is} of a list. Line~2 constrains the depth of the  nested input to two. We handle higher dimensional input separately.
Line~3 applies a predicate that is true when \inline{O} can be unified as the expected output for an input \inline{I},
and Line~4 is a recursive constraint, which intuitively continues with the next inputs.
The rule in Lines~6--9 is similar, except that it handles (layer) inputs with a higher dimension, which is checked in Line~7, and recursively uses the initial predicate from Line~1 since the dense layer only performs computations in the innermost list even when it receives high dimensional input data.
Line~11 (and Line~18) are the base cases for the recursion, i.e., when only an empty list remains.

The predicate in Line~13  encodes the main layer functionality and becomes true when the \inline|Res| variable is the expected output for the input \inline{[I|Is]}.
It has the same arguments as the first rule and an additional temporary variable \inline|Res0| for the result. 
It consists of clauses for multiplying the weight arrays \inline|IW| with each input \inline|I| and for adding the results in Line~16. The predicates \inline|multiply_list_with| and \inline|add_lists| are straightforward and are therefore omitted.
\begin{lstlisting}[language=Prolog,  xleftmargin=15pt, float=tp,
%aboveskip=-6pt,belowskip=-6pt, 
caption={Prolog semantics of the Dense layer~\cite{exais}.},
 label={lst:dense},deletekeywords={is}, morekeywords={dense_layer,dense_node_comp,depth,add_lists,multiply_list_with,padding1D}]
dense_layer([I|Is], IWs, Bs, [O|Os]) :-
    depth([I|Is],2),
    dense_node_comp(I, IWs, Bs, O),
    dense_layer(Is, IWs, Bs, Os).
    
dense_layer([I|Is], IWs, Bs, [O|Os]) :-
    depth([I|Is],D), D > 2,
    dense_layer(I, IWs, Bs, O),
    dense_layer(Is, IWs, Bs, Os).
    
dense_layer([], _, _, []).

dense_node_comp([I|Is],[IW|IWs],Res0,Res) :-
    multiply_list_with(IW,I,Res1),
    add_lists(Res0,Res1,Res2),
    dense_node_comp(Is,IWs,Res2,Res).
    
dense_node_comp([],[],Res,Res).
\end{lstlisting}
With this Prolog semantics, we can now answer a variety of queries, e.g., to compute the expected output of a Dense layer. 

More relevantly, ExAIS contains preconditions that reflect layer requirements, like a specific input shape, or dependencies between the arguments. An example precondition to check if layer input data has a minimum number of dimensions is illustrated in Listing~\ref{lst:precondition}. The predicate 
\inline{check_min_dimensions} takes the input data and a minimum dimension value as arguments. Line~2 shows a predicate that becomes true, when \inline{D1} can be unified to the dimension number of \inline{Is}. Next, there is a condition to check if the dimensions of the input are smaller than the given minimum value. If it is smaller, then 
an error message is produced. Otherwise, the predicate becomes true.
\begin{lstlisting}[language=Prolog,  xleftmargin=15pt, float=tp,
aboveskip=-6pt,belowskip=-6pt, 
caption={Precondition to check if the input data has a minimum number of dimensions.},
 label={lst:precondition},deletekeywords={is}, morekeywords={dense_layer,dense,dense_node_comp,depth,add_lists,multiply_list_with,padding1D,check_min_dimensions,writeln,shape,term_string,string_concat,throw}]
check_min_dimensions(Is, D) :-
	depth(Is,D1),
	(D1 < D ->(write("Invalid Model, Badness Value: "),
			BV is D1-D,BV1 is BV*100000000000000000, 
			writeln(BV1),
			S1 = "Dimension error, Input Shape ",
			shape(Is,Shape),
			term_string(Shape,S2),
			string_concat(S1,S2,RS),
			S3 = ", Expected Min Dimensions ",
			string_concat(S3,D,RS1),
			string_concat(RS,RS1,S),
			throw(S));true).
\end{lstlisting}
Most layers of ExAIS contain preconditions in this form.
\rev{The preconditions are part of the ExAIS semantics and were created manually by the ExAIS developers according to the TensorFlow documentation and according to other publications that describe the layer functionality and requirements.}
When an AI model is executed with the semantics, then all the preconditions of the individual layers are checked. Any violation of the preconditions would make the model invalid.
Hence, the preconditions can help identify problematic model aspects. In this work, we utilize this feature to enable our automated AI model repair approach.
\rev{The execution of the Prolog predicates works similarly to the execution of functions. There is a predicate for each layer, which contains precondition calls and calls to subpredicates, which enables an automatic precondition check during the layer execution. There might be layer preconditions that are not specified in ExAIS and their violations would not be repairable by our approach, but since we tested thousands of random models with our approach, we are confident that we can resolve precondition violations in most cases.
}

The semantics consists of 65 deterministic layers and seven non-deterministic layers. We focus on repairing deterministic layers in this work. 
Only six of the 65 layers have no preconditions (i.e., the Flatten layer, ReLU, ThresholdedReLU, LeakyReLU, Masking, TimeDistributed).
Seven  of the layers have simple preconditions (UpSampling1D-3D, ZeroPadding1D-3D, Embedding), i.e., they only require input with a certain number of dimensions.
The remaining layers have non-trivial preconditions that mainly fall in the following categories. First, there are consistency requirements between the layer arguments (or inputs) and the weight shape, e.g., for a dense layer the first dimension of the weight needs to have the same size as the last dimension of the input.
Second, we may have inconsistencies among the shapes of multiple inputs of a layer, i.e., some layers perform mathematical operations, like addition or multiplication of multiple layer inputs. These layers may require that the shape of the inputs must be the same.
Third, there can be consistency requirements between the layer arguments or with the layer arguments and the inputs. For example, for some convolutional layers setting a dilation\_rate value not equal to one is incompatible with specifying any stride value not equal to one.

\rev{During our investigations, we noticed that there are various studies that evaluate AI faults (based on bug reports) and that some of them are related to violations of the layer preconditions of ExAIS and can thus be detected by its precondition checks.
For example, one study} \cite{DBLP:conf/icse/IslamPNR20} \rev{ investigated bug reports and it illustrates data and layer dimension errors, which make up close to 30\% of the findings. Moreover, the study showed a
broad categorisation of the manual repairs that were suggested in the bug reports.
Another study} \cite{DBLP:conf/icse/HumbatovaJBR0T20} \rev{discusses Tensor shape errors that are related to input shape violations that can be captured with ExAIS's preconditions.
The source of most bugs of those studies was not directly related to a misuse of layers or a wrong network architecture, which made
them inapplicable for our approach, but their frequency highlights the significance of such bugs.

Additionally, we performed experiments with randomly generated models, and collected bug reports from stackoverflow as we will explain in Section}~\ref{sec:evaluation} \rev{ 
and in both cases we observed bugs that are related to precondition violations of ExAIS. 
Hence, we believe that bugs based on the precondition violations are relevant and that it is important to provide better ways to fix such bugs.
}
Based on the three categories of preconditions and the simple dimension preconditions, we developed repair algorithms for the issues that are identified with the preconditions. We explain these algorithms in the following section.

\begin{table*}[tp]
\centering
\caption{Repair approaches for different types of bugs.}\label{tab:fixbugs}
\begin{tabular}{|l|c|c|}
\hline
 & \textbf{Repair Approaches} & \textbf{Examples} \\ \hline
\begin{minipage}[t]{0.14\linewidth}
\vspace{-6pt}
\textbf{Dimension Error}
\end{minipage}       &  \begin{minipage}[t]{0.41\linewidth}
\vspace{-6pt}
\textbf{Input data needs more dimensions}
\begin{itemize}
	\item replacement of the layer that causes the error with a lower dimension version if it is available
	\item insertion of a reshape layer that adds a dimension with size one to increase the number of dimensions of the data
\end{itemize}
\textbf{Input data needs fewer dimensions}
\begin{itemize}
	\item replacement of the layer that causes the error with a higher dimension version if it is available
	\item insertion of a reshape layer that reduces the number of dimensions by combing (multiplying) the last two dimensions into one dimension (e.g., a shape (2,3,2) would become (2,6))
\end{itemize}
\vspace{0.05em}
\end{minipage} &
\begin{minipage}[t]{0.37\linewidth}
\vspace{-6pt}
\textbf{Model with a bug}
\begin{lstlisting}[language=Python, numbers=none, basicstyle=\scriptsize]
model = tf.keras.Sequential([
	Dense(10, input_shape=(10,10,10,)),
	Conv2D(10, kernel_size=(2,2)),
	MaxPooling3D(pool_size=(3,3,3))])
\end{lstlisting}
\textbf{Fixed model}
\begin{lstlisting}[language=Python, numbers=none, basicstyle=\scriptsize]
model = tf.keras.Sequential([
	Dense(16, input_shape=(10,10,10,)),
	Conv2D(16, kernel_size=(2,2)),
	MaxPooling2D(pool_size=(3,3))])
\end{lstlisting}
\end{minipage}
\\ \hline
%
%
\begin{minipage}[t]{0.14\linewidth}
\vspace{-6pt}
\textbf{Input Shape Bug}
\end{minipage}       &
\begin{minipage}[t]{0.41\linewidth}
\vspace{-6pt}
\textbf{Multiple inputs of a layer need to have matching shape or dimension sizes}
\begin{itemize}
	\item addition of a ZeroPadding1D-3D layer to adjust one of the input shapes according to the expected one from the error message
	(if the dimensions of the input are between three and five) 
	\item insertion of Concatenate layer that can adjust the input shape by 
	combining a specified shape with additional values
\end{itemize}
\vspace{0.05em}
\end{minipage}

&
\begin{minipage}[t]{0.37\linewidth}
\vspace{-6pt}
\textbf{Model with a bug}
\begin{lstlisting}[language=Python, numbers=none, basicstyle=\scriptsize]
R1 = ReLU(input_shape=(2,2,)
R2 = ReLU(input_shape=(5,2,)
A  = Add()([R1, R2])
\end{lstlisting}
\textbf{Fixed model}
\begin{lstlisting}[language=Python, numbers=none, basicstyle=\scriptsize]
R1 = ReLU(input_shape=(2,2,)
Z  = ZeroPadding1D(padding=(2,1))(R1)
R2 = ReLU(input_shape=(5,2,)
A  = Add()([Z, R2])
\end{lstlisting}

\end{minipage}
\\\hline \multirow{2}{*}{\vspace{10pt}\textbf{Argument Error}  }      &
\multirow{2}{*}{\begin{minipage}[t]{0.41\linewidth}
\vspace{-6pt}
\textbf{Layer arguments have consistency requirement violations among each other or with the input}
\begin{itemize}
	\item step-wise (random) regeneration of the layer arguments by increasing the number of arguments that are changed
\end{itemize}
\textbf{Weight shape bugs due to weight data that does not match with other arguments/input data or has wrong dimensions}
\begin{itemize}
	\item regeneration of the layer weights while considering the other layer arguments and input shape
\end{itemize}
\textbf{Pool or kernel shapes that are too large for the input shape}
\begin{itemize}
	\item regeneration of the pool/kernel shape
	\item regeneration of layer arguments, like padding 
\end{itemize}
\end{minipage}}
&
\begin{minipage}[t]{0.37\linewidth}
\vspace{-6pt}
\textbf{Layer with a bug}
\begin{lstlisting}[language=Python, numbers=none, basicstyle=\scriptsize]
Cropping1D(input_shape=(4,4,),cropping=(2,2))
\end{lstlisting}
\textbf{Fixed layer}
\begin{lstlisting}[language=Python, numbers=none, basicstyle=\scriptsize]
Cropping1D(input_shape=(4,4,),cropping=(2,1))
\end{lstlisting}
\vspace{3.25em}
\end{minipage}\\
\cline{3-3}
 & &
\begin{minipage}[t]{0.37\linewidth}
\vspace{-6pt}
\textbf{Layer with a bug}
\begin{lstlisting}[language=Python, numbers=none, basicstyle=\scriptsize]
MaxPooling2D(input_shape=(8,8,8,),pool_size=(9,9))
\end{lstlisting}
\textbf{Fixed layer}
\begin{lstlisting}[language=Python, numbers=none, basicstyle=\scriptsize]
MaxPooling2D(input_shape=(8,8,8,),pool_size=(4,3))
\end{lstlisting}
\vspace{3.25em}
\end{minipage}
  \\ \hline
\end{tabular}
\end{table*}

\section{Method} \label{sec:method}
In this section, we describe how we fix specific bugs and how our repair algorithm for AI models works in detail.

\rev{We consider a repair to be valid if it removes TensorFlow errors (that occur during the model execution) with minimal adjustments.
We created our repair suggestions with the intention to preserve the original model as much as
possible, i.e., we were looking for minimal model adjustments that maintain the layers and most of the structure of the model. The motivation behind this is the fact that developers usually make
small mistakes} \cite{DBLP:journals/corr/abs-2104-02517}. \rev{For simple cases, like dimension errors, our repairs follow the best practice, as suggested by the accepted solutions from bug reports. For more complicated cases, like argument or shape issues, we developed repairs that only utilise standard layers} \cite{layerdocs}
\rev{
(or argument modifications). There would be various other repair options if more deep learning operations, like NumPy functions} \cite{numpy}, \rev{
would be considered. We believe that fixes with standard layers are well suited since they are straightforward, easier to understand, and many AI developers stick to these layers.
Generally, it is hard to justify the quality
of more complicated repairs without an additional training and model validation step. We intend to further explore these steps in future work.
Overall, we believe that our repairs are reasonable, especially since they often were equivalent to accepted solutions for real bugs as we will explain later in Section}~\ref{sec:evaluation}.

To illustrate the repair of specific model bugs, we present our approaches for fixing common errors that are related to the misuse of standard deep learning layers, i.e., dimension, input shape, and argument errors. 
Bugs that originate from other AI development tasks, like the data preprocessing, training, or model validation, are out of the scope of this work.
Table~\ref{tab:fixbugs} gives a simplified overview of our repair approaches for these errors and shows example fixes.

For dimension errors, there are two cases: the expected number of dimensions is larger than the actual number of dimensions, or the opposite.
For the first case, we check if the problematic layer can be replaced with a higher dimensional version (e.g., Conv2D with Conv3D). 
Alternatively, this case can be repaired by adding a reshape layer with dimension size one before the layer with the bug. A reshape layer can modify the shape of the input data, while it still keeps the same number of values, i.e., the product of the dimension sizes will be the same after a reshape.
The second case, i.e., the expected number of dimensions is smaller than the actual dimension number, can be handled similarly. A layer can be replaced with a smaller dimensional version, but in contrast to the previous case, the reshape works differently. In order to obtain a smaller dimension number, a new shape is computed by combining the last two dimensions of the given input (by multiplying them).
Both these repair options are considered for our algorithm. 
The final minimal suggested fix is determined based on which of the two fixes leads to a smaller overall change.

Next, for input shape bugs, there are also two potential repairs. 
Inconsistent input shapes can occur when a layer that takes multiple inputs (e.g., Add) has incompatible inputs with different shapes or dimension sizes.  
One way to resolve this bug, is to add padding around one of the inputs, i.e., with a ZeroPadding layer that increases the input space and adds zeros around the given data. An example repair with such a layer is illustrated in the second row of Table~\ref{tab:fixbugs}. It shows a graph model with an Add layer that has two ReLU layers with different shapes as input.
The fixed model has an additional ZeroPadding1D layer that is in-between the ReLU layer with the smaller input and the Add layer.
A padding layer can resolve most of such bugs, but not all, since there are only padding layers that take three to five-dimensional inputs. 
Alternatively, if the dimension is outside this range, the shape mismatch 
can be fixed with a Concatenate layer. This layer can adjust the input shape by combining a certain dimension of the input with additional values or arrays of values, which enables our repair approach to change the input shape according to the requirements of the invalid layer.

Thirdly, for argument errors, there are also a number of possible repairs.
Most of these errors can be repaired by regenerating the layer arguments, i.e., by randomly exchanging the arguments of the layer with random values. 
This procedure works step by step, first a replacement of a single argument is tried and if this is not successful then more arguments are exchanged.
A special case of an argument error is a weight shape bug that occurs when the weight data does not match with other arguments/input data of the layer or when it has a wrong dimension number. This case can be repaired by regenerating the weights of the layer while considering the other layer arguments and the input shape. 

Another case that needs to be handled separately are pool (or kernel) shapes that are inconsistent with the input shape, because they are too large. 
These bugs can be fixed by adopting the pool (or kernel) shapes or by regenerating other layer arguments, like padding.
There are more special cases like this that have their own preconditions and need some specific approaches to be repaired, which were omitted for brevity.
There are two example layers given for this bug type as illustrated in the last row of Table~\ref{tab:fixbugs}. First, there is a Cropping1D layer, that applies too much cropping, which would result in an empty output. The bug is fixed by reducing the cropping size values.

\begin{algorithm*}[t]
	\caption{Pseudo code of the repair algorithm.}
	\label{alg:repair}
	\small
	\begin{algorithmic}[1]
		\Require
		$\mathit{SemanticHelper}$ helper class for the execution of the semantics 
		
		\Function{findFixWithMinimalChange}{rand,model, error, maxFixes}
		\Comment{find single fix with minimal change in the AI model}
		\State $\mathit{workingFixes\gets \{ \}}$
		\State $\mathit{fixingCount \gets 0}$
		\Do
		\State $\mathit{fixes \gets errorSpecificFixes(rand, model, error, maxFixes)}$ \Comment{produces fixes based on the approaches of Table~\ref{tab:fixbugs}}
		\ForEach {$fix \in Fixes $}
		\State $\mathit{(success, newError) \gets SemanticHelper.run(fix)}$\Comment{run semantics, get success/error}
		\If{$\neg success \land \mathit{(newError.getBadness() < error.getBadness() \lor newError.getLocation() \geq error.getLocation())}$}
		\State $\mathit{fix \gets findFixWithMinimalChange(rand,fix,newError,maxFixes)}$
		\EndIf
		\State $\mathit{workingFixes.add(fix)}$
		\EndFor%
		\State $\mathit{fixingCount \gets fixingCount + 1}$
		\doWhile{$workingFixes.isEmpty() \land fixingCount < maxFixes$}
		\Return $getFixWithSmallestChangeValue(workingFixes)$ \Comment{returns the fix with the smallest change value}
		\EndFunction
		\newline
		\Function{findFixes}{rand, model, error, maxFixes}
		\State $\mathit{workingFixes \gets \{\}}$
		\State $\mathit{fixes \gets errorSpecificFixes(rand, model, error, maxFixes)}$\Comment{produces fixes based on the approaches of Table~\ref{tab:fixbugs}}
		\ForEach {$fix \in Fixes $}
		\State $\mathit{(success, error) \gets SemanticHelper.run(fix)}$\Comment{run the semantics, get success/error}
		\If{$\neg success$}
		\State $\mathit{fix \gets findFixWithMinimalChange(rand,fix,error,maxFixes)}$
		\Comment{uses the first function}
		\EndIf
		\State $\mathit{workingFixes.add(fix)}$
		\EndFor%
		\Return $\mathit{sortByChangeValue(workingFixes)}$\Comment{sorts the fixes accoring to their change value}
		\EndFunction
	\end{algorithmic}
\end{algorithm*}

Another example shows a pool shape bug in a MaxPooling2D layer that has a pool shape that is too large for the specified input shape. A fix for this bug is produced by replacing the pool sizes with smaller values.
It should be noticed that in reality such bugs are much harder to find since the
models are larger, have many arguments, and the input and output shapes of layers are often not easy to see.
Moreover, a model can usually not just be repaired with a singular change of a layer since in many cases there need to be further adjustments deeper in the neural network. For example, 
argument regenerations often change the output shape of a layer, which can
cause inconsistencies with the weight or shape requirements in the next layers. The following simple example model illustrates this cascading behaviour.
\begin{lstlisting}[language=Python, numbers=none, basicstyle=\scriptsize]
C1 = Conv1D(2,input_shape=(8,1,),kernel_size=2,dilation_rate=3)
C2 = Conv1D(2,input_shape=(16,1,),kernel_size=2,dilation_rate=3,strides=3)
S  = Subtract()([C1, C2])
\end{lstlisting}
It shows a Subtract layer that has two Conv1D layers as inputs. The model seems to be valid at a first glance. Both Conv1D layers would produce the same output shape, since the stride argument (that specifies the step size with which the kernel is moved over input data) offsets the larger input shape of the second convolution.
However, a stride value greater than one is incompatible with a dilation\_rate value greater than one. (A dilation\_rate can be specified to expand a kernel with zero values.)
A regeneration of the arguments to fix this violation, e.g., by replacing the stride or dilation\_rate value, will always change the output shape.
As a consequence, there will be an inconsistent input shape error at the Subtract layer, which needs to be fixed as shown in Table~\ref{tab:fixbugs}.

The overall repair approach that incorporates the specific fixes (from Table~\ref{tab:fixbugs}) is outlined in 
Algorithm~\ref{alg:repair}. It consists of two major functions. 
A function that tries to find a singular working fix with a minimal change compared to the original model, and a function that returns a number of potential working fixes by applying the first function.
The first function $\mathit{findFixWithMinimalChange}$ takes a random object, a model to be fixed, an error object, and a maximum number of fixes that should be considered as input. 
In Lines~1--2, we initialise a set $\mathit{workingFixes}$ and a counter $\mathit{fixingCount}$. Then, there is a do-while loop that continues until there is a working fix, or until the maximum number of allowed fixing attempts is reached. Within the loop, the $\mathit{errorSpecificFixes}$ function (Line~5) is called to receive the potential fixes for a given error. 
For each of the fixes, we apply our $\mathit{SemanticHelper}$ to check if it produced a valid model. The $\mathit{SemanticHelper}$ is a wrapper class that 
helps with the execution of ExAIS to make a prediction for a given model, by returning a success message or an error object.
If it returns an error, then we check if the error helped to improve the model, i.e., with the help of a badness value (Line 8), 
and try to further repair the model by recursively calling the function.
The badness value is calculated by preconditions of ExAIS and returned with an error message, and it works similar as fitness in search-based software testing~\cite{DBLP:conf/laser/HarmanMSY10}.
It is a distance metric that is larger when the layer arguments and inputs are far from being valid, i.e., the layer preconditions are ranked based on severity and a value is calculated by taking into account the difference of an observed and an expected argument value and by multiplying it with a severity factor \cite{exais}.

Finally, in Line~13, a helper function is used 
that sorts our working fixes based on a change value that indicates the similarity to the original model, and return the fix with the smallest change.
The change value is another distance metric that is calculated by considering the number of layer arguments, the layer replacements and additions that are required to repair the original model. We consider an argument modification the smallest change (change value 1), followed by a layer replacement (change value 5), and the largest changes are layer additions (change value 10).

The second function $\mathit{findFixes}$ has the same arguments, and also initializes a set for the fixes.
Line~16 applies the $\mathit{errorSpecificFixes}$ function to obtain potential fixes for the specified error. Then, for each of these fixes, we check if it is working with the $\mathit{SemanticHelper}$. If it is not, then we apply the $\mathit{findFixWithMinimalChange}$ to recursively find a working fix. The fix is added to the fix set, which is sorted and returned at the end of the function. The difference of this function to the first one is that it produces 
a number of potential fixes instead of a single one.

\begin{lstlisting}[language=Python,  xleftmargin=15pt, float=tp, 
	%aboveskip=-5pt,belowskip=-7pt, 
	caption={Real model with a bug from stackoverflow \cite{bug1}.}, label={lst:modelbug}]
	import tensorflow as tf
	from tensorflow.keras import layers, models
	model_valid = tf.keras.Sequential([
	layers.Flatten(input_shape=(10,)),
	layers.Dense(16,  activation='relu'),
	layers.Conv1D(16, kernel_size=(2), activation='relu', padding='same'),
	layers.MaxPooling1D(pool_size=(4), strides=3, padding='valid'),
	layers.Flatten(),
	layers.Dense(1, activation='softmax')
	])
\end{lstlisting}
\begin{lstlisting}[language=Prolog,  xleftmargin=15pt, float=*,
	%aboveskip=-6pt,belowskip=-6pt, 
	caption={ExAIS Prolog code of a neural network with a bug.},
	label={lst:prologbug},deletekeywords={is}, morekeywords={dense_layer,flatten_layer,conv1D_layer,max_pool1D_layer,exec_layers}]
	LFla71197 = flatten_layer([[1.2079020849786344, 1.449234775683509, ...]], Fla71197), 
	LDen79959 = dense_layer(Fla71197, [[0.6039479770827674, 0.410518536377564, ...]], Den79959), 
	LCon60545 = conv1D_layer(Den79959, 2,[[[0.6711,...]]],[...], 1, true, 1, Con60545), 
	LMax46787 = max_pool1D_layer(Con60545, 4, 3, false, Max46787), 
	LFla21960 = flatten_layer(Max46787, Fla21960), 
	LDen25740 = dense_layer(Fla21960, [[0.6182226038831956, ...]],[...], Den25740), 
	exec_layers([LFla71197,LDen79959,LCon60545,LMax46787,LFla21960,LDen25740],["Fla71197","Den79959","Con60545","Max46787","Fla21960","Den25740"],Den25740,"Den25740").
\end{lstlisting}
\begin{lstlisting}[language=Python,  xleftmargin=15pt, float=tp, 
	%aboveskip=-5pt,belowskip=-7pt, 
	caption={Potential repair for the buggy model.}, label={lst:fix}]
	import tensorflow as tf
	from tensorflow.keras import layers, models
	model_valid = tf.keras.Sequential([
	layers.Flatten(input_shape=(10,)),
	layers.Dense(16,  activation='relu'),
	layers.Reshape((16,1)),
	layers.Conv1D(16, kernel_size=(2), activation='relu', padding='same'),
	layers.MaxPooling1D(pool_size=(4), strides=3, padding='valid'),
	layers.Flatten(),
	layers.Dense(1, activation='softmax')
	])
\end{lstlisting}

Listing~\ref{lst:modelbug} illustrates a real neural network in the form of a TensorFlow Python program that was posted on stackoverflow \cite{bug1} and contains a bug. The model is invalid, because the Conv1D layer requires input data with three dimensions, but it receives a two-dimensional input.
In order to run such a neural network with ExAIS, it needs to be converted into a Prolog model, which is shown in Listing~\ref{lst:prologbug}. 
\rev{Fortunately, this conversion can be done automatically with our repair tool. Our tool supports models in an JSON format which is based on a format introduced by SOCRATES}~\cite{socrates}. \rev{AI models can be exported into this format in a straightforward way. We automatically convert these JSON models into Prolog models (in the form of queries).}
The model consists of predicate calls for the layers that are assigned to variables (Lines 1--6).
For the execution of the model (i.e., making a prediction) there is an  \inline|exec_layers| predicate in Line~7. This predicate takes a list of layer variables, executes them, checks the preconditions and helps to identify the name of a layer that violates a precondition. 
Since the model is invalid, it cannot be used for a prediction. The execution returns the following error message to show the precondition violation.
\begin{lstlisting}[language=Prolog,  numbers=none, %xleftmargin=20pt, 
 morekeywords={dense_layer,sig_gate,depth,add_lists,multiply_list,tanh_gate}]
Invalid Model, Badness Value: -100000000000000000
Aborted at Con60545: Dimension Error, Input Shape [1,16], Expected Dimensions 3!!!
\end{lstlisting}
This message gives us the necessary information to repair the bug, i.e., it shows the problematic layer and what needs to be changed.
A fix that was produced for the error is shown in Listing~\ref{lst:fix}. 
This fix was computed by the $\mathit{findFixWithMinimalChange}$ function  and it has minimal changes, i.e, it is the most similar to the original model, since the only change is the insertion of a reshape layer. 
The reshape adds a dimension with size one to the input data to satisfy the dimension requirements for the Conv1D layer.
Alternatively, such an error could be resolved by looking for a lower dimensional version of the Conv1D layer, but since it is already the version with the lowest number of dimensions, this is not feasible in this case.

\section{Evaluation} \label{sec:evaluation}
In this section, we evaluate the effectiveness and performance of our approach.
We implemented our approach that can support various types of neural networks and tested it with a variety of AI models. To demonstrate its usefulness, we show two example applications, i.e., repairing automatically generated neural networks, and repairing manually designed neural networks with bugs.
We perform multiple experiments to answer the following research questions (RQ).

\begin{itemize}
	\item  \emph{RQ1: How effective is our approach in repairing automatically generated models?}
	There are various issues, like inconsistencies with layer arguments, or a faulty network structure, that can result in invalid neural networks. It is important to evaluate to what extent we can repair such bugs.
	
	\item \emph{RQ2: How effective is our approach in repairing manually written models?}
	Developing an AI model is an error-prone task. To highlight the usefulness of our repair approach in practice, we demonstrate the applicability of our approach for a number of real modelling bugs, and investigate to what extent our suggested repairs are relevant for fixing these bugs.

	\item \emph{RQ3:  What kind of AI models can we support with our approach, and how efficient is our repair method?}
	To clarify the scope of our supported models, it is important to explain what kind of layers and model types we support. Moreover, it also makes sense to investigate up to what network size our approach still runs in a reasonable time.
\end{itemize}
The experiments were performed on a 7th Gen. Lenovo X1 Carbon ThinkPad with an 8th Gen i7 CPU with four 1.80 GHz cores and 16 GB RAM. 
For executing the Prolog semantics, we used SWI-Prolog 8.2.1, and our repair tool was built in Java 13.0.7. It consists of about 10,000 lines of code and includes some of the functionality of the ExAIS test case generator in order to produce Prolog models. Moreover, it uses the JSON.simple library and 
can load and save models in a JSON format that is based on the format from SOCRATES~\cite{socrates}. Additionally, it can store models as TensorFlow programs, and visualize them with Dot from Graphviz.
\rev{The tool together with our experiment data and results are available in our repository} \cite{reponew}. 
In the following, we present our answers to the research questions.

\paragraph{RQ1: How effective is our approach in repairing automatically generated models?}
There are a number of approaches that automatically generate AI models for the purpose of AI framework testing.
Wang et al.~\cite{DBLP:conf/sigsoft/WangYCLZ20} show a differential testing approach called LEMON that uses mutations to generate AI models. The approach produces 100\% valid models since the mutations are designed to generate only working models, but LEMON is limited to 24 types of layers.

A fuzzing method that extracts information from AI library documentations was illustrated by Li~\cite{li2020documentation}.
The approach generates AI models based on learned input requirements of deep learning layers, but it only produces about 25\% valid models that are limited to singular layer models.

Another fuzzing approach that generates AI models was presented by Schumi and Sun~\cite{exais}. The method is semantic-based and produces AI models with an optimisation algorithm with some rudimentary fix components that work more random and focus more on replacing problematic layers. 
\rev{The approach focuses more on generating valid models, but incorporates rudimentary repairs, for dimension errors and input shape bugs. The algorithm works by adding, replacing or deleting layers and it is guided by a badness value that indicates how far the model is away from being valid. However, it restarts
the generation if it cannot find a working model or it just deletes problematic model components. Hence, in many cases it does not perform a fix.}
It is able to produce 99.1\% valid models for differential testing.
We use this approach as a baseline to compare it to our method.

In order to evaluate the effectiveness of our approach, we performed an experiment with 1,000 randomly generated invalid AI models. The models were produced with an adopted generation approach from \cite{exais} by randomly connecting various types of layers. Many layers 
are connected sequentially by feeding the output as input to the next layer in the sequence. However, there are also layers that take multiple inputs, e.g., to perform mathematical operations, like addition. For these layers, multiple input layers are connected. It should be noted that a random model that is produced in such a way is nearly always invalid due to various layer preconditions that must be satisfied.

The resulting neural networks had an average size of 7.4 layers (excluding input layers). We set the parameters to use small inputs with a range of one to four values per dimension, \rev{since bigger sizes vastly increased the overall size of the highly dimensional input data (with up to 7 dimensions) and slowed down the execution of ExAIS.} 
On average there were 1.8 bugs per model and in total there were 1797 bugs that needed to be fixed. An overview of the types of bugs is shown in Table~\ref{tab:modelbugs}. It can be seen that argument errors are the most common, followed by input shape bugs and dimension errors.
Our approach successfully repaired all the bugs, i.e., \rev{we were able to repair 100\% of the random bugs that occurred within the randomly generated neural networks}, and it took on average only 21.08s per model.
\rev{Being able to repair such a wide set of models with various different layer types shows that our approach is highly effective.}
In order to compare our approach to the baseline, we executed the same 1,000 invalid random tests with the optimisation algorithm from~\cite{exais}. The approach was able to find a valid fix that was similar to our solution in 87.9\% of the models, i.e., the layers of the model were kept and similar adoptions as with our algorithm were made. \rev{It took on average 8.96s, which is about twice as fast as our approach, 
because it discards models (or model parts) early when they are not working, and since it does not include a comprehensive search algorithm that tries to find minimal working fixes.}
For 121 cases, it did not produce a model \rev{ that we would consider a fix since the algorithm was randomly replacing or removing layers or regenerating the whole model, i.e., the fixed models were altered in a way that made them no longer recognisable compared to the original model.}
Hence, we believe that our method showed a significant improvement of this state-of-the-art approach.

\begin{table}[tp]
	\centering
	\caption{Overview of the produced model bugs.}
	\label{tab:modelbugs}
	\begin{tabular}{|l|c|c|c|c|}
		\hline
		\textbf{Type of Bug} & \textbf{Number of Bugs} \\ \hline
		Dimension Error      & 219 \\ \hline
		Weight Shape Issue    & 45 \\ \hline
		Input Shape Bug       & 705 \\ \hline
		Argument Error        & 828  \\ \hline
		\textbf{Total}        & 1797 \\ \hline
	\end{tabular}
\end{table}

\paragraph{RQ2: How effective is our approach in repairing manually written models?}
In order to answer this research question, we collected a set of real modelling bugs from AI developers that were reported on stackoverflow. We searched 
issues for which people were struggling to find a solution for an error message from TensorFlow that was caused by a misuse of one of the standard layers or a wrong combination of these layers. 
To find these issues, we used search queries based on common error messages from TensorFlow that we obtained with the previous random test models.
\rev{For example, we searched with error specific queries, like ``TensorFlow shape must be rank'', ``TensorFlow inconsistent input shapes'', ``Operands could not be broadcast together with shapes'', and with general queries, like ``TensorFlow layer argument error'' or ``TensorFlow layer dimension error''. 
We found hundreds of bugs reports that we had 
to manually inspect  since the vast majority of them had causes unrelated to  the model or layer definition, i.e., they were not relevant to our approach.}

Some of the bugs are outside the scope of our method, since the cause of these bugs is not related to the model structure or layers. AI development has several steps, like the data preprocessing, training, or model validation, where bugs can occur, but our approach is only concerned with bugs within the models, e.g., by a wrong use of standard layers. Moreover, many models contain hand-crafted layers or enhanced features that are not supported by the semantics.

We found 16 bugs that matched our criteria, five with dimension errors \cite{bug1,bug2,bug3,bug4,bug9}, three with invalid kernel or pool shapes \cite{bug5,bug6,bug7}, five weight shape issues \cite{bug10,bug13,bug14,bug15,bug16}, and three bugs with inconsistent arguments or input shapes \cite{bug8,bug11,bug12}. An overview of these bugs, their type, model sizes and repair times is shown in Table~\ref{tab:realbugs}.
For 15 of the models, our method was able to find potential repairs. One model \cite{bug3} was too large for the execution with the semantics since the vocabulary data (of an Embedding layer) was too much (25001x200) for Prolog to handle, but after reducing the vocabulary size (to 2501x200), which was irrelevant to the bug in the model, our method was able to produce a repair. 
On average the repair time was 3min 40s, the maximum time was 23min 21s. The model sizes ranged from about 1KB up to 43MB.

The results show the effectiveness of our method in repairing real modelling bugs, which we believe is important because it seems to be common that AI developers have difficulties to correctly design an AI model or to understand error messages from AI frameworks. 

\begin{table*}[t]
\centering
\caption{Overview of the real model bugs and their types and model sizes.}
\label{tab:realbugs}
\begin{tabular}{|l|c|c|c|c|c|c|c|}
\hline
& \textbf{References} & 
\textbf{\begin{tabular}[c]{@{}c@{}}Number\\ of Bugs\end{tabular}} & 
\textbf{\begin{tabular}[c]{@{}c@{}}Min Model\\ Size\end{tabular}} & 
\textbf{\begin{tabular}[c]{@{}c@{}}Max Model\\ Size\end{tabular}} & 
\textbf{\begin{tabular}[c]{@{}c@{}}Min Repair\\ Time[s]\end{tabular}}& 
\textbf{\begin{tabular}[c]{@{}c@{}}Max Repair\\ Time[s]\end{tabular}}&  \textbf{\begin{tabular}[c]{@{}c@{}}Avg. Repair\\ Time[s]\end{tabular}}  \\ \hline
\textbf{Dimension Error}       & \cite{bug1,bug2,bug3,bug4,bug9} & 5 & 4KB   & 10MB  & 5 & 301 & 69\\ \hline
\textbf{Pool/Kernel Shape Bug} & \cite{bug5,bug6,bug7}           & 3 & 35KB  & 6MB  & 5 & 501 & 181 \\ \hline
\textbf{Weight Shape Issue}&\cite{bug10,bug13,bug14,bug15,bug16} & 5 & 1KB   & 43MB  & 5 & 1402 & 388\\ \hline
\textbf{Input Shape Bug}       & \cite{bug11,bug12}              & 2 & 2MB   & 5MB   & 238 & 254 & 246 \\ \hline
\textbf{Argument Error}        & \cite{bug8}                     & 1 & 186KB & 186KB & 217 & 217 & 217 \\ \hline
\textbf{Total}        & -                     & 16 & 1KB & 43MB & 5 & 1402 & 220 \\ \hline
\end{tabular}
\end{table*}

\begin{table*}[t]
\centering
\caption{Comparison of the repairs of the real model bugs to the solutions from stackoverflow.}
\label{tab:comparebugs}
\begin{tabular}{|l|c|c|c|c|}
\hline
 &
  \textbf{\begin{tabular}[c]{@{}c@{}}Equivalent \\ or Closely Related\end{tabular}} &
  \textbf{\begin{tabular}[c]{@{}c@{}}Comparable\\ Quality\end{tabular}} &
  \textbf{\begin{tabular}[c]{@{}c@{}}Not Comparable\\ but Still Valid\end{tabular}} & \textbf{Total} \\ \hline
\textbf{Dimension Error}       & 4 & 1 & - & 5 \\ \hline
\textbf{Pool/Kernel Shape Bug} & - & 1 & 2 & 3 \\ \hline
\textbf{Weight Shape Issue}    & 4 & 1 & - & 5 \\ \hline
\textbf{Input Shape Bug}       & - & 2 & - & 2 \\ \hline
\textbf{Argument Error}        & - & - & 1 & 1 \\ \hline
\end{tabular}
\end{table*}

%

In order to further evaluate the practical usefulness of our approach, we evaluated our produced repairs for these manual bugs by comparing them to the accepted solutions from stackoverflow. 
Table~\ref{tab:comparebugs} gives an overview of this comparison. Out of the 16 models, eight of our repairs were equivalent or very close (in terms of the type and magnitude of the modification) to the suggested solutions on stackoverflow \cite{bug1,bug3,bug4,bug9,bug10,bug13,bug14,bug15}. For example, for dimension issues, our solutions would be to introduce resize layers as it was also suggested on stackoverflow.
Five of our repairs were of similar quality compared to the posted solutions \cite{bug2,bug7,bug11,bug12,bug16}, e.g., for inconsistent weight shapes we usually regenerate the layer weights based on the other layer arguments.
Alternatively, such issues can be addressed by adjusting the layer arguments to correct the weight shape.
Our remaining three repairs were not related to the suggested stackoverflow solutions \cite{bug5,bug6,bug8}, but they still produced working models that might be helpful as an alternative solution for the bugs.
For example, a wrong pool/kernel shape can be corrected in multiple ways. Our solution to randomly regenerate the layer arguments step by step usually produced a valid repair, but there can still be other better solutions.
Based on these results, we believe that our repairs are reasonably useful in practice and can be helpful for various real modelling bugs.

\paragraph{RQ3:  What kind of AI models can we support with our approach, and how efficient is our repair method?}
Since our approach is based on the existing semantics ExAIS, we support neural networks that are supported by the semantics. ExAIS supports 65 deterministic layers 
of various types, like convolutional, pooling, recurrent, activation, normalisation, mathematical, and cropping. It was developed for TensorFlow and support almost all of its layers, i.e., the developers built ExAIS for TensorFlow 2.4 and identified 72 unique non-abstract and non-wrapper layers. Only 7 of these layers were not included. 
A full list of the supported layers is available in the repository of ExAIS \cite{reponew}. Hence, our repair approach is able to repair a variety of neural networks, and nearly all types of layers.

In order to evaluate to which extent our approach works with different model sizes and how fast it can repair these models, we set up two experiments. (1) We evaluate the repair time of randomly generated neural networks of different types with an increasing number of layers. For this generation, we restricted the types of layers since the execution time of the semantics is very different depending on the type of layers.
(2) We use models for benchmark datasets with more realistic input and weight data sizes in order to evaluate the performance in a practicable setting.

\begin{figure}[!t]
	\centering
	\begin{tikzpicture}[scale=0.98]
	\begin{axis}[,
	try min ticks=5,
	legend style={at={(0.0,0.28)},anchor=south west},
	width=1.0\linewidth,
	height=12em,
	xticklabel style={/pgf/number format/precision=6,/pgf/number format/fixed},
	scaled ticks=false,
	ymajorgrids,
	ylabel={run time [s]},
	xlabel={model size (number of layers)},
	]
	\addplot table[x=size,y=duration] {runtimeDenseG.csv};
	\addlegendentry{dense};
	\addplot table[x=size,y=duration] {runtimeRecuG.csv};
	\addlegendentry{recurrent};
	\addplot table[x=size,y=duration] {runtimePoolG.csv};
	\addlegendentry{pooling};
	\addplot table[x=size,y=duration] {runtimeConvG.csv};
	\addlegendentry{conv};
	\end{axis}
	\end{tikzpicture}
	\caption{Run time of our repair algorithm for graph models.} 
	\label{fig:runtime1}
\end{figure}
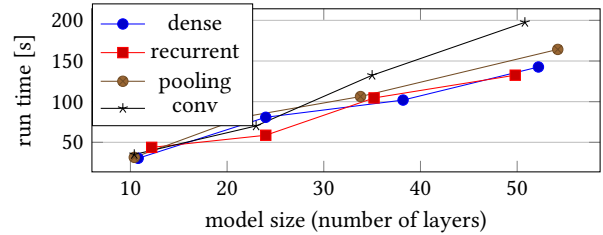

The results of the first experiment are shown in Fig.~\ref{fig:runtime} and Fig.~\ref{fig:runtime1}, which illustrate the repair time of our approach for different types of models and for increasing layer numbers. Fig.~\ref{fig:runtime} shows sequential models that include activation layers and (1) dense layers, (2) recurrent layers, (3) pooling layers, and (4) convolutional layers. Fig.~\ref{fig:runtime1} shows the same, but also includes graph models that include forks as a result of mathematical layers with multiple inputs.
It can be seen that even relatively large models can be repaired in a reasonable time. Sequential models with up to 50 layers can be repaired in 80s, even if they contain mostly complex layers like convolutional or pooling. 
Graph models with about 50 layers take longer, but can still be repaired in max 3min 40s.
We believe that compared to the manual effort of inspecting and repairing such a large model, our repairing time is reasonably practical.


For the second experiment, we applied three models for the well-known CIFAR~\cite{cifar} and MNIST datasets~\cite{mnist} that were provided by SOCRATES~ \cite{socrates}.
The models were called \lstinline[]{cifar_conv_small_relu}, \lstinline[]{mnist_conv_small_relu_diffai} 
and \lstinline[]{mnist_conv_9_200}.
Additionally, we used a model~\cite{fashionmodel} for the Fashion-MNIST dataset~\cite{fashionmnist} and another model~\cite{svhnmodel} for the Street View House Numbers (SVHN) dataset~\cite{svhn}.
In contrast to the previously generated models, these neural networks had input data of up to 64KB and a model size of up to 10MB due to bigger weight size. 
In order to evaluate our approach for these models, we manually added inconsistent arguments, dimension errors, and weight shape issues.
On average, it took 3min 12s to repair these models, and the maximum repair time was 5min 54s. 
This demonstrates that we are also able to apply our approach to practical neural networks and not just randomly generated ones. Moreover, it shows that the repair time of these larger neural networks was still reasonable, especially since the experiments were only performed on a laptop with rather limited computing resources and without parallelisation.

\begin{figure}[!t]
	\centering
	\begin{tikzpicture}[scale=0.98]
	\begin{axis}[,
	try min ticks=5,
	legend style={at={(0.0,0.28)},anchor=south west},
	width=1.0\linewidth,
	height=12em,
	xticklabel style={/pgf/number format/precision=6,/pgf/number format/fixed},
	scaled ticks=false,
	ymajorgrids,
	ylabel={run time [s]},
	xlabel={model size (number of layers)},
	]
	\addplot table[x=size,y=durationDenseSeq] {runtime.csv};
	\addlegendentry{dense};
	\addplot table[x=size,y=durationRecuSeq] {runtime.csv};
	\addlegendentry{recurrent};
	\addplot table[x=size,y=durationPoolSeq] {runtime.csv};
	\addlegendentry{pooling};
	\addplot table[x=size,y=durationConvSeq] {runtime.csv};
	\addlegendentry{conv};
	\end{axis}
	\end{tikzpicture}
	\caption{Run time of our repair algorithm.} 
	\label{fig:runtime}
\end{figure}
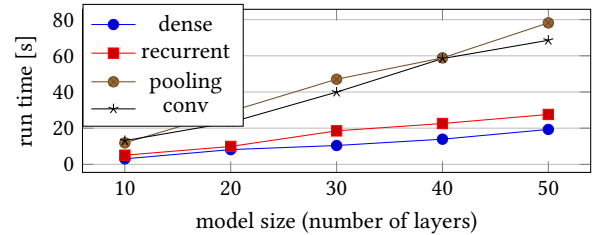

\paragraph{Discussion.}
A potential threat to the validity of our evaluation might be that we focused too much on models with small input sizes. 
Generally, AI models can handle huge data like voice recordings, or large images.
An evaluation with larger test inputs and models might be more realistic, but a large input size (in the MB range) will produce even bigger models, which would soon cause memory overflows in the semantics.
However, usually bugs in large models can be broken down to smaller version that are easier to process.
We believe that our test models with rather small inputs were still reasonable and did not represent a big limitation, and it is well-known that small test cases can reveal various bugs~\cite{DBLP:conf/issta/JacksonD96}. Moreover, we tested reasonably sized models from several benchmark datasets, and we evaluated real model bugs. 

One might argue that the performance of our approach for larger models is  limited, since it can take a couple of minutes to execute our approach for large models. We believe that the performance is still reasonable especially since it can avoid a lot of effort that would be required to manually inspect a model for bugs, or when the long resolution time for AI bug reports is considered that can range from weeks to months \cite{DBLP:conf/issre/ThungWLJ12}.

Another threat to the validity of our evaluation might be a potential bias when we selected the real model bugs. It is true that we had quite restricted criteria when we were looking for these bugs. 
There are numerous bug reports from AI developers that are outside the scope of this work since they are not related to the misuse or wrong composition of standard layers. Moreover, there are still many  bugs that come from AI models that include non-standard features, like custom layers, which make them unusable with our approach.
Hence, it was a cumbersome task to look for relevant bugs for our approach. 
We believe that we still managed to find a representative set of model bugs that was suited for a good evaluation of our method. There might be rare bugs in the same category that are not supported by our method, but our evaluation showed that we are able to repair common modelling bugs in a reasonable time.

Another question that might come up is why we do not use error messages from an AI framework instead of utilizing a semantics to support our repair approach? It is true that our algorithms would in principle also work with error messages from AI frameworks and it would even be faster, but there are a number of problems with these messages \cite{exais}. The error messages are not always consistent, i.e., even for the same type of bug there can be various different messages, which might be caused by independent implementations in different layers.
The source of an error is not always clear and the necessary debug information to repair an error can be hard to extract since the messages have no clear and consistent form.
In contrast, ExAIS provides error messages that are well-structured and consistent. They can easily be automatically parsed and provide clear debug information and the source of a bug.
Moreover, it is easy to extend ExAIS with additional preconditions, which can be helpful for checking custom model properties or it enables the identification of problematic behaviour that might not lead to error messages in an AI framework. Hence, we believe that it was a good choice to apply ExAIS for our repair approach.
\section{Related Work}\label{sec:related}%
Most of the related work in neural network repair focuses on other aspects of the neural networks, like improving the accuracy or ensuring certain properties. 

For example, Sohn et al.~\cite{DBLP:journals/corr/abs-1912-12463} introduce a 
repair technique called Arachne that can improve pre-trained models by adjusting
the weights of layers. The method applies differential evolution to optimise the weights and to correct misbehaviour, like misclassifications. Moreover, they demonstrate how to resolve fairness issues, i.e., by repairing a bias in a gender classifier.

Another approach that deals with fairness properties was presented by Sun et al.~\cite{sun2022causality}. The work illustrates a causality-based repair technique called CARE that can identify problematic neurons that are responsible for undesired neural network behaviour. It is able to ensure that a neural network satisfies various fairness and safety properties, e.g., it can remove backdoors caused by malicious training data.

Yang et al.~\cite{DBLP:conf/formats/YangYTHJP22} shows a repair framework that is able to ensure safety and robustness with regard to input-output safety specifications. They illustrate a depth-first-search reachability analysis algorithm to find unsafe input regions and examples that represent these regions. The approach is evaluated with an aircraft collision avoidance and a rocket landing system. Similarly, Sotoudeh and Thakur~\cite{DBLP:conf/pldi/SotoudehT21} present a provable point repair algorithm that is able to deal with misclassifications and that can ensure safety properties.

Usman et al.~\cite{DBLP:conf/cav/UsmanGSNP21} illustrates a constraint-based  repair method called NNREPAIR for neural network classifiers. The technique applies fault localization to find problematic network parameters, and it can improve the accuracy of a model and fix safety properties.

Xie et al.~\cite{DBLP:conf/icml/Xie0MLWZLX21} introduced a model-based repair approach for recurrent neural networks (RNN) called RNNRepair. It is based on an influence model that relates the behaviour of a network to the training data. The method can help to understand the behaviours of an RNN, as well as increase the accuracy and repair safety properties. 

\rev{
Zhang et al.}~\cite{DBLP:conf/icse/ZhangZMS21} \rev{introduced a DNN training monitoring and automatic repairing tool called AUTOTRAINER that can fix common training problems, like a slow convergence or fluctuating accuracies.}
\rev{Similarly, a tool  called DeepDiagnosis that further improved the repair performance for such bugs, was introduced by Wardat et al.}~\cite{DBLP:conf/icse/WardatCLR22} \rev{It applies dynamic analysis to monitoring and detect errors according to various symptoms.  
It can fix eight different training problems and can do this more efficient and with a better performance than other tools.
In contrast to both these methods, our approach is only concerned with the model and layer definitions and not with the training phase.}

Related work is also in the field of AI model debugging which includes a number of approaches and tools~\cite{DBLP:journals/tvcg/StrobeltGBPPR19,DBLP:conf/icml/OdenaOAG19,DBLP:conf/chi/SchoopHH21,DBLP:journals/tvcg/HohmanKPC19,DBLP:conf/sigsoft/MaLLZG18} that offer features like fuzzing, faulty neuron (or feature) localisation, visualisation, or auditing, to enable finding bugs that lead to inaccuracy or property violations. 
However, such approaches are usually limited to specific types of models, e.g., classifiers.

In contrast to all these approaches, our work is not concerned with the accuracy or with certain properties about the predictions of a neural network.
We focus on repairing modelling bugs that cause invalid neural networks with structural problems, wrongly used layers, or inconsistent layer arguments.

\rev{The closest related work is the tool called Tensfa from
Wu et al.~}\cite{DBLP:conf/issre/WuSC0Q21,DBLP:journals/infsof/WuSCJQ22}. \rev{
The tool can automatically detect and repair tensor shape faults, which are comparable to our dimension and input shape bugs. For these bugs, the approach 
works for even broader scope of AI models and programs since it supports more general operations with Tensors, like array adjustments with NumPy} \cite{numpy}, \rev{ which are not standard layers. The approach uses decision tree model to detect bugs based on crash messages from TensorFlow. It applies static data dependence analysis and a dynamic shape tracking techniques to locate faults, and multiple mutation strategies that perform repairs by considering the frequency and time cost of a number of repair patterns. Some of the repair patterns, like the introduction of reshape layers, are similar to our fixes, but since Tensfa uses more operations than just standard layers, it has more repair options for fixing shape issues. Additionally, Tensfa can check if the input and output shapes of a model are causing a bug, which is out of the scope of our method.

In contrast, our repair method with ExAIS works more straight forward. To identify and localize faults, we just run a model with ExAIS and the precondition checks will identify and localize the bugs. 
Moreover, our approach is not just limited to Tensor shape bugs, since we support other issues, like argument or weight shape errors, and the focus of our repairs is on finding minimal changes with standard layers.
Lastly, Tensfa can only fix issues when there is an error message. ExAIS can also identify rare bugs that do not produce error messages from TensorFlow}~\cite{issue6}.

To the best of our knowledge, our work is the first that shows an automatic semantic-based repair approach for AI model bugs.
\section{Conclusion} \label{sec:conclusion}
We have introduced a novel neural network repair approach that is based on an executable semantics and demonstrated two applications. Our method is able to repair invalid AI models that suffer from structural problems, wrongly used layers, or wrongly connected layers.
It works by executing a given AI model with an existing AI framework semantics called ExAIS that has built-in preconditions that are able to provide error messages with debug information that helps to localise and characterise a bug.
Based on these error messages that are produced when there is a precondition violation, we are able to 
identify and repair invalid model aspects with a number of algorithms for different types of bugs.

One major application of our approach is the repair of automatically generated neural networks that can, e.g, be used for AI framework testing.
In order to evaluate the effectiveness of our approach for this application, we generated 1,000 random models with bugs and repaired them.
Our approach was able to repair 100\% of these neural networks, and it took on average only 21.08s to completely repair the models. Moreover, we evaluated the approach by repairing a set of larger test models (for well-known benchmark datasets) which required about 192s on average. 

Another application that we presented is the repair of real modelling bugs from AI developers. For this use case, we collected a set of neural network bugs that developers were struggling with from stackoverflow. Out of 16 faulty AI models, we could directly repair 15 models, and also the one remaining model could be repaired after a minor size adjustment of the model.  The average repair time for the bugs was only 3min 40s. 
Inspecting the quality of our repairs, showed that 13 of our produced repairs were equally as good or were comparable to the solutions on stackoverflow.

To sum up, our approach was able to effectively and automatically produce practical repairs for real world bugs within minutes. This shows that it can be valuable for AI developers since it can reduce a lot of debugging effort. 

We believe that these two approaches highlight the usefulness and applicability of our approach, which has potential to enable further applications.
In the future, we aim to explore semantic-based repair techniques for other AI model aspects.

\emph{Acknowledgments.}
This research is supported by the Ministry of Education, Singapore under its Academic Research Fund Tier~3 (Award ID: MOET32020-0004). Any opinions, findings and conclusions or recommendations
expressed in this material are those of the author(s) and do not reflect the views of the Ministry of Education, Singapore.

\onecolumn
\begin{multicols}{2}
\bibliographystyle{ACM-Reference-Format}
\bibliography{main}
\end{multicols}
\end{document}